%% file: main.tex
\documentclass[10pt,twocolumn,letterpaper]{article}

\usepackage[pagenumbers]{cvpr}
\input{preamble}
\usepackage{fancyhdr}

\begin{document}

\title{Score Normalization for Demographic Fairness in Face Recognition}


\author{\vspace*{-.5ex}\hspace*{-.7em}Yu Linghu$^1$\,
Tiago de Freitas Pereira$^3$\,
Christophe Ecabert$^2$\,
S\'ebastien Marcel$^2$\,
Manuel G\"unther$^1$\\[1ex]
\hspace*{-3em}
\begin{minipage}[t]{.36\textwidth}\small
$^1$Department of Informatics\\
\hspace*{.5em}University of Zurich\\
\hspace*{.5em}Andreasstrasse 15, CH-8050 Zurich \\
\hspace*{.5em}\texttt{\{yu.linghu,manuel.guenther\}@uzh.ch}\\
\hspace*{.5em}\texttt{https://www.ifi.uzh.ch/en/aiml.html}
\end{minipage}
\begin{minipage}[t]{.29\textwidth}\small
$^2$Idiap Research Institute\\
\hspace*{.5em}Centre du Parc, Rue Marconi 19 \\
\hspace*{.5em}CH-1920 Martigny \\
\hspace*{.5em}\texttt{\{cecabert,marcel\}@idiap.ch}\\
\hspace*{.5em}\texttt{https://www.idiap.ch}
\end{minipage}
\begin{minipage}[t]{.3\textwidth}\small
$^3$ams OSRAM\\
\hspace*{.5em}\texttt{tiago.defreitaspereira@ams-osram.com}\\
\hspace*{.5em}\texttt{https://ams-osram.com/} \\
\end{minipage}\vspace*{-.5ex}
}

\maketitle

{
  \chead{\footnotesize This is a pre-print of the original paper accepted for presentation at the International Joint Conference on Biometrics (IJCB) 2024.}
  \lhead{}
  \thispagestyle{fancy}
}

\input{texfiles/abstract}
\input{texfiles/intro}
\input{texfiles/related}
\input{texfiles/approach}
\input{texfiles/evaluation}

\input{texfiles/experiments}
\input{texfiles/discussion}
\input{texfiles/conclusion}

\section*{Acknowledgement}
The authors thank the Hasler foundation for
their support through the SAFER project.

{
\small
\bibliographystyle{ieee}
\bibliography{texfiles/Publications,texfiles/References}
}

\end{document}

%% file: preamble.tex
\usepackage{times}
\usepackage{graphicx}
\usepackage[norule,symbol,perpage]{footmisc}

\usepackage{epsfig}
\usepackage{lmodern}
\usepackage{multirow,tabularx}
\usepackage[point]{fltpoint}
\usepackage{textcomp}
\usepackage{amsmath,amssymb}
\usepackage{ifthen}
\usepackage{color}
\usepackage{xcolor}
\usepackage{tabularx}
\usepackage{booktabs}
\usepackage{afterpage}

\usepackage[listofformat=parens, justification=centering, subrefformat=subparens,font=footnotesize,skip=3pt]{subfig}

\usepackage[breaklinks=true,colorlinks,bookmarks=false]{hyperref}

\DeclareGraphicsExtensions{.pdf,.jpg,.png}
\graphicspath{{./graphics/}{./figures/}{./results/}}

\newcolumntype{C}{X<{\centering}}

\definecolor{lightgreen}{rgb}{0.67, 0.88, 0.69}
\definecolor{darkgreen}{rgb}{0,0.55,0}
\definecolor{linkcolor}{rgb}{0,0,.65}

\newcommand\ig[2][1]{\includegraphics[width=#1\textwidth, page=1]{#2}}

\newcommand\Caption[3][]{\caption[#2]{\label{#1}\textsc{#2}. \small#3}}
\renewcommand\etal[1]{\textit{et al.}~\cite{#1}}

\renewcommand\sec[1]{Sec.~\ref{sec:#1}}
\newcommand\fig[1]{Fig.~\ref{fig:#1}}

\newcommand\tab[1]{Tab.~\ref{tab:#1}}
\newcommand\stab[1]{Tab.~\subref{tab:#1}}

\DeclareMathOperator*{\argmax}{arg\,max}



%% file: texfiles/abstract.tex
\begin{abstract}
Fair biometric algorithms have similar verification performance across different demographic groups given a single decision threshold.
Unfortunately, for state-of-the-art face recognition networks, score distributions differ between demographics.
Contrary to work that tries to align those distributions by extra training or fine-tuning, we solely focus on score post-processing methods.
As proved, well-known sample-centered score normalization techniques, Z-norm and T-norm, do not improve fairness for high-security operating points.
Thus, we extend the standard Z/T-norm to integrate demographic information in normalization.
Additionally, we investigate several possibilities to incorporate cohort similarities for both genuine and impostor pairs per demographic to improve fairness across different operating points.
We run experiments on two datasets with different demographics (gender and ethnicity) and show that our techniques generally improve the overall fairness of five state-of-the-art pre-trained face recognition networks, without downgrading verification performance.
We also indicate that an equal contribution of False Match Rate (FMR) and False Non-Match Rate (FNMR) in fairness evaluation is required for the highest gains.
Code and protocols are available.\footnote[9]{\url{https://github.com/AIML-IfI/score-norm-fairness}\label{fn:github}}

\end{abstract}

%% file: texfiles/intro.tex
\section{Introduction}
\label{sec:intro}

\begin{figure}[t]
  \ig[.45]{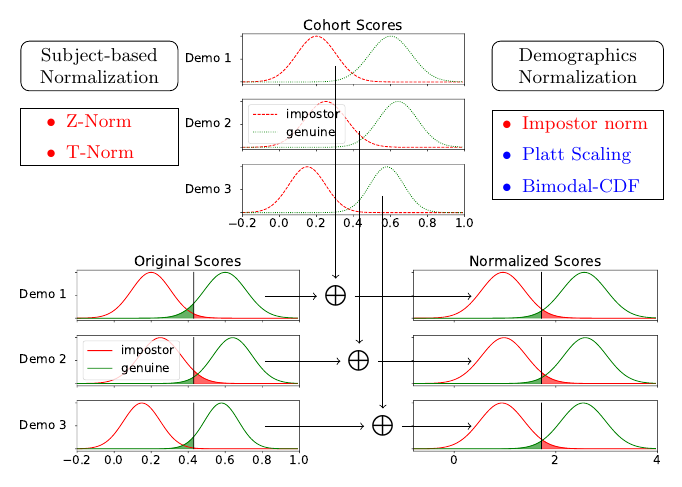}
  \Caption[fig:teaser]{Improved Fairness Through Score Normalization}{
    The original scores on the left have different False Match Rates (FMR, red area) and False Non-Match Rates (FNMR, green area) for different demographics under the same score threshold.
    Through modeling of score distributions from a cohort, we normalize scores such that they provide more similar FMR and FNMR across demographics, thereby improving demographic fairness.
    Normalization techniques in red text use cohort impostor scores only, blue ones also incorporate cohort genuine scores.}
\end{figure}

The automatic identification and verification of facial images gained large attention in the last decades.
With the advent of deep learning, many new methods \cite{liu2017sphereface,wang2018cosface,deng2019arcface,meng2021magface,kim2022adaface} and facial image datasets \cite{zheng2018cross,cao2018vggface2,maze2018ijbc,wang2019racial,zhu2021webface260m} have been developed to train and evaluate deep learning methods.
These methods have matured into being usable in security-relevant applications like automatic border control using e-gates \cite{delrio2016automated}.

The applicability of these algorithms highly depends on the characteristics of the demographic groups in which they are employed.
It was observed that \emph{The Other Race Effect}, which is well-known in humans \cite{malpass1969recognition}, can also be observed in Face Recognition (FR) algorithms. 
Since most large-scale datasets include mainly images from white people \cite{wang2019racial}, and dataset biases are learned by deep learning algorithms \cite{rudd2016moon,albiero2020sexist,albiero2020balance}, research has shown that algorithms perform very well on white male populations, but decrease performance on females and/or people of color \cite{albiero2020gender,krishnapriya2020skintone}.
Consequently, most news media coverage that reports the wrong behavior of automatic FR algorithms finds the higher false negative rate in the latter demographics \cite{romm2018amazons,hill2020wrongfully}.
Therefore, the Face Recognition Vendor Test (FRVT) has a special report addressing demographic effects in FR \cite{grother2019frvt}, mostly observing the effect of ethnicity and gender on more than 100 Commercial-Off-The-Shelf (COTS) systems.

In face verification, a similarity value is computed between a previously enrolled gallery template (such as the face image stored in a passport) and a probe image, \eg taken in an e-gate.
A threshold is applied to this similarity value to decide whether the gallery template and the probe image come from the same identity.
One major problem with fairness in biometrics is that the distributions of similarity scores differ between demographics, for example, the mated comparison of African people usually results in lower similarity scores than White \cite{vangara2019characterizing}.
Hence, a single threshold can have differential performance across demographics \cite{howard2019effect,pereira2021fairness}.


Our proposed approaches to overcome this issue include score normalization techniques that have been successfully applied in FR \cite{wallace2012cross,mandasari2014calibration}. 
As shown in \fig{teaser}, the advantage of these techniques is that they can be applied to any existing FR system to improve fairness across different people and do not require any further network training.
In one of these techniques, Z-norm \cite{reynolds2000speaker}, the score is normalized by a gallery-sample-specific distribution of similarities to cohort samples of different subjects.
This method has been tested in \cite{kotwal2024mitigating}.
In this paper, we modify these techniques to perform score normalization for a certain demographic group -- instead of per sample -- and show that the impostor score distribution can be normalized with this technique, but the genuine score distributions will be more disparate.
Therefore, we test methods that normalize both genuine and impostor score distributions at the same time.
We investigate Platt scaling \cite{platt1999probabilistic} and propose a new method.
We show that most score normalization techniques can improve demographic fairness by a good margin, by experimenting on two different datasets with a total of six evaluation protocols and five different pre-trained deep networks.

As our contributions in this paper, we:
\begin{itemize}\vspace*{-1ex}
  \setlength\itemsep{-.3em}
  \item propose score normalization methods at post-processing stage for bias mitigation without downgrading verification performance;
  \item extend Z/T-norm to integrate demographics and propose three cohort-based methods, with one fitting the distribution for impostor scores and the other two also considering genuine scores;
  \item develop a new protocol for the RFW dataset with impostor scores selected randomly and define cohorts for the original and our new protocol;
  \item examine all methods on gender and ethnicity with feature extractors that perform differently;
  \item and analyze the relative contribution of FNMR and FMR in fairness evaluation.
\end{itemize}

%% file: texfiles/related.tex
\section{Background and Related Work}
\label{sec:related}

In recent years, deep learning has dominated and revolutionized the field of FR.
Two main research directions exist in the biometrics community: developing better network topologies and implementing better-suited loss functions.
Wang \etal{wang2021survey} provide a survey of algorithms, datasets, and evaluations before 2018, and more approaches have been developed since.
Most modern network architectures \cite{hu2018squeeze} include variations and improvements \cite{duta2021iresnet} of residual network architectures \cite{he2016deep}.
The latest developed loss functions, \ie{} ArcFace \cite{deng2019arcface}, MagFace \cite{meng2021magface}, and Adaface \cite{kim2022adaface}, improve the discriminability of deep features in angular space.

Also, aspects of fairness have been discussed and evaluated \cite{grother2019frvt,wang2019racial}.
Cavazos \etal{cavazos2020accuracy} describe underlying factors that bias COTS FR systems with respect to ethnicity.
It was observed that biases are more frequently observed in low-quality samples \cite{grother2019frvt,cavazos2020accuracy}.
Vangara \etal{vangara2019characterizing} show consistently higher False Match Rates (FMR) in African Americans compared to Caucasians using several COTS systems.

In most cases, face verification performance is evaluated via two error measures \cite{phillips2011evaluation}.
For a given score threshold, the FMR computes the number of impostor comparisons (gallery template and probe sample are from different identities) with a similarity above the threshold.
Similarly, the False Non-Match Rate (FNMR) calculates the number of genuine pairs (two samples of the same identity) with a similarity below the threshold.
Oftentimes, these rates are computed and averaged across demographics.
Since most datasets have mainly Caucasian people \cite{wang2019racial}, errors made on non-Caucasians do not influence results much.

Calculating True Match Rates (TMR) for each demographic label individually is widely used to evaluate the fairness differences across demographics \cite{huang2019deep,kortylewski2019analyzing,qin2020asymmetric,wang2020mitigating,wang2019racial,wang2019deep}.
Though one can find differences in the performances of different algorithms \cite{krishnapriya2020skintone}, it is common to set a single threshold and analyze differences across demographic groups since independent thresholds are rarely applied in production.
Some works in the biometrics literature advocate that the threshold should be demographic-specific \cite{cook2019demographic,poh2010group}.
On the other hand, Fairness Discrepancy Rate \cite{pereira2021fairness}, Inequality Rate \cite{grother2021demographic} and Gini Aggregation Rate for Biometric Equitability \cite{howard2022evaluating} integrate within-demographic FMR$(\tau)$ and FNMR$(\tau)$ differences to measure the bias.
Other methods consider the FNMR differences \cite{schuckers2022statistical} and score distribution differences \cite{kotwal2022fairness} across demographics.
Here we use the metric suggested by the National Institute for Standards and Technology (NIST) \cite{grother2022nist} which compares the Worst-case Error Rate to the geometric mean of FMR and FNMR.

There also exist methods to mitigate the bias in FR systems by improving features extracted from the networks \cite{gong2020jointly,liang2019additive,morales2020sensitivenets,gong2021mitigating,huang2019deep,qin2020asymmetric,serna2022sensitive,xu2021consistent,yang2021ramface,sukthanker2022importance,yang2022enhancing}, partially solving ethical problems by generating synthetic images for different demographics \cite{atzori2023demographic, rahimi2023toward}.
With features extracted via existing FR systems, Terhorst \etal{terhoerst2020comparison} train a classifier to replace the regular cosine similarity function which pushes score distributions of different groups to be similar.
A fair score normalization (FSN) method proposed by Terhorst \etal{terhorst2020post} uses KMeans to cluster features and combine the cluster-specific thresholds and global thresholds to normalize scores.
Kotwal \etal{kotwal2024mitigating} propose a score calibration method to align the score distribution by fine-tuning a pre-trained network with additional intra- and inter-demographic loss terms.
In contrast, we solely focus on boosting fairness of existing FR systems without any network training process and no need to access features.
While somewhat related to our work (Kotwal \etal{kotwal2024mitigating} investigate similar methods as ourselves, such as M1 and M2, see below), they made use of a network pre-trained on MS1MV3 \cite{guo2016msceleb1m}, which overlaps identities with the RFW test set \cite{wang2019racial}.

%% file: texfiles/approach.tex
\section{Approach}
\label{sec:approach}

\input{results/method.tex}

To apply a single score threshold $\tau$ that is suited for different demographics, it is required that each demographic follows a similar score distribution.
We investigate several techniques that provide this capability, a list of these techniques is provided in \tab{method}.
Most techniques collect score distributions $D=\{\mathrm{sim}(g_1,p_1), \mathrm{sim}(g_2,p_2), \ldots\}$ which are computed by choosing various gallery $g
$ and probe samples $p$.
These distributions are modeled to follow a normal distribution $\mathcal N_{\mu,\sigma}$ with mean $\mu$ and standard deviation $\sigma$:
\begin{equation}
  \label{eq:parameters}
  \begin{aligned}
    \mu & = \mathbb E\bigl(D\bigr) & \sigma &= \sqrt{\mathbb E\bigl((D - \mu)^2\bigr)}
  \end{aligned}
\end{equation}
Finally, standardization is performed for a given similarity score $s$ between gallery $g$ and probe sample $p$ from the test set:
\begin{equation}
  \label{eq:score-norm}
  \begin{aligned}
    s' = \mathcal S_{\mu, \sigma}(s) &= \frac{s-\mu}{\sigma} & \text{with}\quad s &= \mathrm{sim}(g,p)
  \end{aligned}
\end{equation}
Methods using only impostor scores to model $\mathcal N_{\mu,\sigma}$ are introduced in \sec{impostor1}, \ref{sec:impostor2}, and \ref{sec:impostor3}, and methods including genuine scores in \sec{both_sides}.

\subsection{Identity-based Score Normalization}
\label{sec:impostor1}
Well-known techniques such as Z-norm \cite{reynolds2000speaker} and T-norm \cite{auckenthaler2004score} fight differences in score distributions of single individuals that were first observed by Doddington \etal{doddington1998sheep}.
The identity-based impostor score distributions are computed by comparing the test gallery $g$ or probe sample $p$ to samples from different identities selected from a cohort dataset.
When the test gallery $g$ is compared to the cohort, this technique is called Z-norm \cite{reynolds2000speaker}, which we refer to as M1.
T-norm \cite{auckenthaler2004score}, \ie{} when the test probe sample $p$ is compared to the cohort, is called M2 in our evaluation.

\subsection{Demographics-based Score Normalization}
\label{sec:impostor2}
While these identity-based approaches can make distributions more similar across demographics, no such guarantee is given and there is no restriction on the demographics of the cohort samples.
In fact, the distribution contains cross-demographic comparisons, which might not reflect realistic impostor attacks.
Hence, an easy extension of the Z-norm and T-norm would be to restrict the samples in the cohort to the same demographic as gallery/probe sample, respectively.
We term these methods M1.1 for Z-norm and M2.1 for T-norm.

When further splitting up the cohort into demographics as done above, the number of scores for distribution estimation is reduced, especially at the tails of very low FMRs, the estimation might not be relevant any longer and downgrade the verification performance.
To increase number of scores to model normal distributions \eqref{eq:parameters}, we can exploit all in-demographics gallery-cohort comparisons for all enrolled gallery samples, to arrive at M1.2.
Similarly, we can utilize all in-demographics probe-cohort comparisons for all probe samples, to arrive at M2.2.
While combining comparisons over all gallery or probe samples can be achieved when working on specific datasets, in typical verification applications there exists no large gallery, and we have access to a single probe only.
Therefore, methods M1.2 and M2.2 are more of a theoretical nature.

\subsection{Pure Cohort-based Score Normalization}
\label{sec:impostor3}
For a more fair and realistic evaluation, we propose to move away from subject-based normalization such as T-norm and Z-norm.
Instead, we solely rely on in-cohort in-demographics comparisons to estimate the score distributions.
Particularly, we select impostor pairs of cohort samples from the same demographics (ethnicity or gender), but from different subjects, and provide mean $\mu_i$ and standard deviation $\sigma_i$ for each demographic $d_i$ via \eqref{eq:parameters}.
We term this method M3.

This way of selecting distributions has two main advantages.
First, these statistics can be pre-computed and do not require additional gallery-cohort or probe-cohort comparisons during enrollment or probing.
Second, the number of scores that can be used is much larger since all different-identity same-demographics pairs are utilized.
The only disadvantage is that we need to know $d_i$ of the comparison to select correct model $\mathcal S_{\mu_i, \sigma_i}$ for \eqref{eq:score-norm}.
Here, we assumed $d_i$ can be gathered during enrollment, and there is no need to know the demographic information for probe sample.

\subsection{Genuine and Impostor Cohorts}
\label{sec:both_sides}
Any of the above approaches have the issue that distributions are only modeled from one type of score, \ie, the impostor scores.
While such methods will likely be able to improve the alignment of the impostor score distributions across demographics, it is unlikely that they also normalize across genuine score distributions.
Thus, they are unlikely to improve algorithmic fairness across all different operating points.

To make use of genuine score distributions, we need to compute similarities for in-cohort pairs with matching identities, which we split into different demographics.
For demographic $d_i$, we mark the genuine score distributions $D_i^\oplus$, while impostor score distributions (which are the same as used in M3) are marked as $D_i^\ominus$.
These two score distributions can now be used to provide a single monotonically increasing function to transform the raw scores into normalized scores.

There exist several techniques to incorporate the set of two distributions $D_i^\oplus$ and $D_i^\ominus$ for a given demographic $d_i$ into one final normalization \cite{poh2005f,poh2009adaptive}.
Our selected representative M4 of these techniques is Platt scaling \cite{platt1999probabilistic,mandasari2014calibration}, where logistic regression is performed to distinguish low impostor scores from large genuine scores by maximizing the weighted binary cross entropy using weights $w^\oplus$ and $w^\ominus$ for normalizing different counts of genuine and impostor scores.
The final logistic function $\sigma$ can be used to normalize the original test score $s$ for a given demographic $d_i$:
\begin{gather}
  \label{eq:platt}
    s' = \sigma(s) = \frac1{1+e^{-\alpha s - \beta}} \qquad\text{with}\\
  \alpha, \beta = \argmax_{\alpha,\beta} \left[w^\oplus\sum\limits_{s \in D_i^\oplus} \log \sigma (s) - w^\ominus \sum\limits_{s\in D_i^\ominus} \log \sigma (s)\right] \nonumber
\end{gather}

Finally, we propose M5 which is related to M3 but incorporates both genuine and impostor score normalization.
This method is inspired by the Bayesian Intrapersonal/Extrapersonal Classifier \cite{moghaddam1998beyond,guenther2009maximumlikelihood}.
We try to estimate a score that combines the probability of being a genuine and not being an impostor score: $\mathcal P(s) = \mathcal P^\oplus(s) - \mathcal P^\ominus(s)$.
The former can be estimated by the Cumulative Distribution Function (CDF) of the genuine score distribution: $\mathcal P^\oplus(s) = \mathrm{CDF}\bigl(\mathcal N^\oplus\bigr)(s)$.
The latter is computed by inverting the CDF of the impostor score distribution: $\mathcal P^\ominus(s) = 1-\mathrm{CDF}\bigl(\mathcal N^\ominus\bigr)(s)$.
By combining and applying them to a specific demographic $d_i$, we arrive at:\\[-2ex]
\begin{equation}
  \label{eq:bimodal-cdf}
  s' = \mathcal P(s) = \mathrm{CDF}\bigl(\mathcal N_i^\oplus\bigr)(s) - 1 + \mathrm{CDF}\bigl(\mathcal N_i^\ominus\bigr)(s)
\end{equation}

%% file: results/method.tex
\begin{table}[!t]
    \footnotesize
    \centering
    \Caption[tab:method]{Score Normalization}{This table lists the score normalization techniques utilized in our experiments, including the data pairs used to compute the statistics.}

    \resizebox{0.93\linewidth}{!}{%
    \begin{tabular}{l|r|r}\toprule
    \bf Method & \bf Description & \bf Data for Statistics \\\midrule
    M1 &Z-norm subject-based &\multirow{3}{*}{Gallery $\times$ Cohort} \\
    \ M1.1  &Z-norm subject-demo-based & \\
    \ M1.2  &Z-norm demo-based & \\[1ex]
    M2 &T-norm subject-based &\multirow{3}{*}{Probe $\times$ Cohort} \\
    \ M2.1 &T-norm subject-demo-based &\\
    \ M2.2 &T-norm demo-based & \\[1ex]
    M3 &Impostor Norm &\multirow{3}{*}{Cohort $\times$ Cohort}\\
    M4 &Platt Scaling & \\
    M5 &Bimodal CDF & \\
    \end{tabular}%
    }
\end{table}

%% file: texfiles/evaluation.tex
\section{Evaluation}

For fairness evaluation, we rely on Worst-case Error Rate to geometric mean of FMR and FNMR (WERM):
\begin{equation}
  \label{eq:werm}
  \begin{aligned}
  \mathrm{WERM}_\tau &= \left(\frac{\max_{d_i} \text{FMR}_{d_i}(\tau)}{\Bigl(\prod_{d_i} \text{FMR}_{d_i} + \epsilon\Bigr)^{1/n}}\right)^{\alpha}\\&\times\left(\frac{\max_{d_i} \text{FNMR}_{d_i}(\tau)}{\Bigl(\prod_{d_i} \text{FNMR}_{d_i} + \epsilon\Bigr)^{1/n}}\right)^{(1-\alpha)}
  \end{aligned}
\end{equation}

Here we use $\epsilon=10^{-5}$.
WERM \cite{grother2022nist} ranges in $(0,\infty)$ where lower values are better.
Often, the contribution of FMR and FNMR are expected to be balanced, but other weights are also possible and should be considered depending on the application needs \cite{pereira2021fairness,howard2022evaluating}.
We apply equal weights: $\alpha=\frac12$.
Since the report on WERM is threshold-specific, we only focus on FMR threshold $\tau = 10^{-3}$ here, and other thresholds can be considered for different application purposes.
Additionally, TMR$(\tau)$ is used to measure verification performance.

\subsection{New Protocols for VGGFace2}
By containing over 3.31 million images of 9131 subjects, the VGGFace2 \cite{cao2018vggface2} dataset is one of the larger FR datasets.
The training set contains 8631 identities, while the test set contains 500 identities.
We utilize the given gender labels (\texttt{Male}, \texttt{Female}), and publicly available ethnicity labels (\texttt{Asian}, \texttt{Black}, \texttt{Indian}, \texttt{White}).\footnote{\url{https://gitlab.idiap.ch/bob/bob.bio.face/-/blob/master/src/bob/bio/face/database/vgg2.py}}
Less than 30 people are removed due to difficulties in determining ethnicities.
Instead of using all samples for the FR experiment, we create a sub-sampled protocol.
For each subject, we randomly pick five samples to compose a probe set and one sample to be registered in the gallery.
We use a sample-to-sample demographic-specific all-vs-all comparison to compute baseline scores $s$.
Since we make use of a cohort to apply the score normalization techniques described above, the same procedure is applied to obtain a cohort from the training set of VGGFace2.

VGGFace2 has a bias in \texttt{White} and \texttt{Male}, which is also reflected in the number of comparisons in the scores.
Since $\tau$ is picked based on impostor score distribution, balancing for demographic groups alters $\tau$.
We want to examine the system fairness variation in the above two situations.
To create balanced numbers of comparisons, for each network, we sub-sample 5,200 impostor pairs per ethnicity, and 180,000 impostor pairs per gender, such that they follow the original distributions of all scores.
As a result, VGGFace2 forms four subsets: VGG2 gender, VGG2 gender-balanced, VGG2 ethnicity, and VGG2 ethnicity-balanced.
The interaction of gender and ethnicity labels is not tested in our experiments, since balancing both at the same time provides splits with very few samples.

\subsection{New Protocols for RFW}
\begin{figure}[t]
  \centering
  \includegraphics[page=1,width=.45\textwidth,clip]{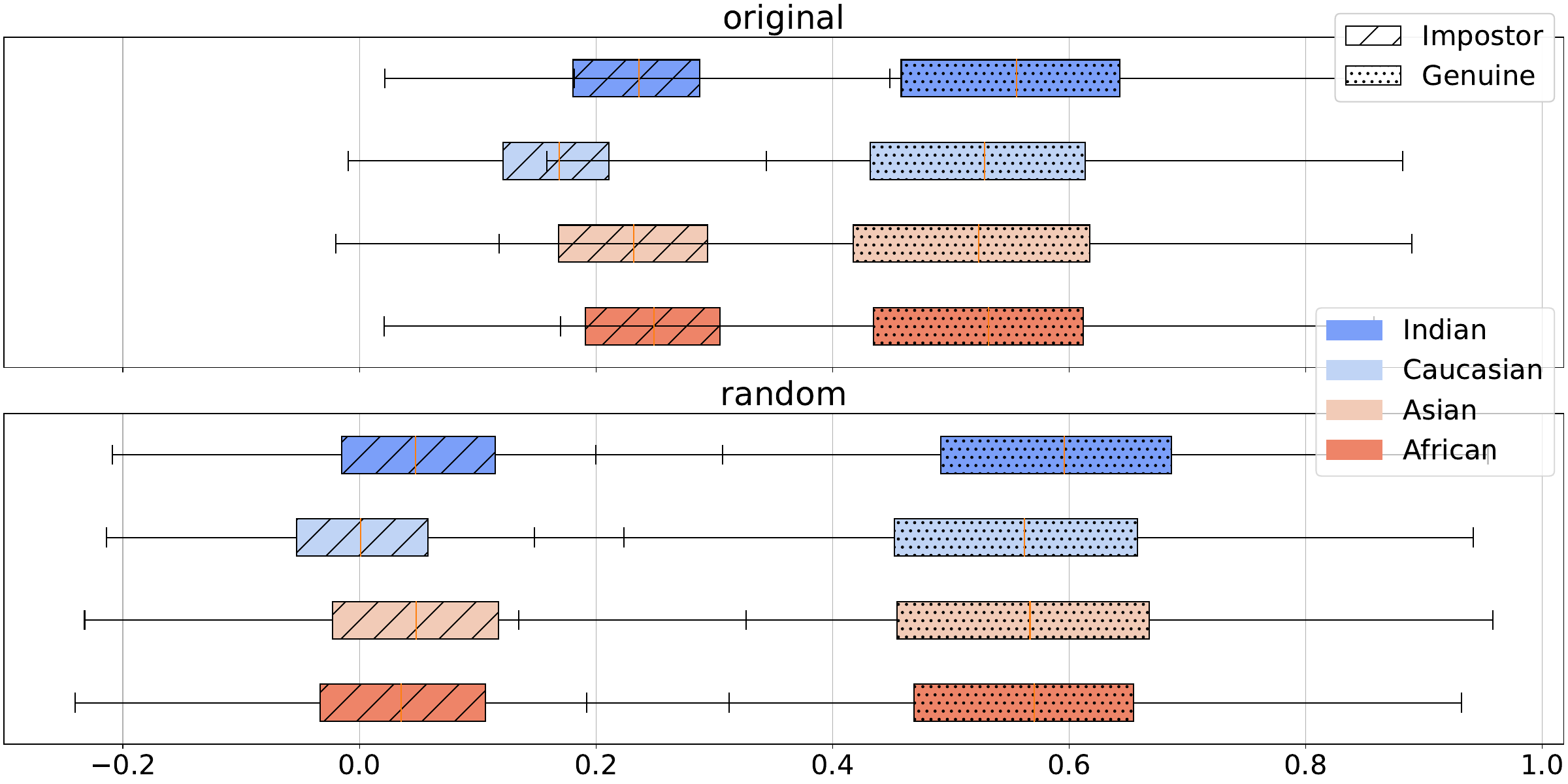}
  \Caption[fig:rfwprotocol]{RFW Protocol Comparison}{
      This figure displays distributions of baseline genuine and impostor scores of the four ethnicities on \texttt{original} and our \texttt{random} RFW protocol computed with the E2 network (cf.~\tab{network}).
  }
\end{figure}
To overcome the issue of unbalanced evaluations with respect to ethnicity, Wang \etal{wang2019racial} introduced the Racial Faces in the Wild (RFW) dataset.
This dataset is a subset of the MS-Celeb-1M (MS1M) \cite{guo2016msceleb1m} whose identities are organized into four different ethnicities (\texttt{African}, \texttt{Asian}, \texttt{Caucasian}, \texttt{Indian}) with about 3000 individuals per ethnicity.
The \texttt{original} verification protocol \cite{wang2019racial} defines around 6000 image pairs (half genuine and half impostor) per ethnicity, utilizing the most similar impostor pairs determined by a deep learning-based FR algorithm, which usually share the same gender and ethnicity.
Thus, the distribution of impostor scores shown in \fig{rfwprotocol} does not follow the general trend of gathering around 0.
Any technique trying to learn this score distribution from cohort data is doomed to fail.
Therefore, we generated a new \texttt{random} protocol for the RFW dataset to make this more comparable with other datasets' results and avoids a possible selection bias in the default protocol.
It is composed of random image pairs (impostor and genuine) of same ethnicity and gender.
The number of pairs is almost identical to protocol \texttt{original}.

To evaluate the difficulty, two protocols are passed into the FR experiment with network E2 (cf.~\sec{experiments}).
At $\tau=10^{-3}$,\footnote{For brevity, we write $\tau=10^{-3}$ to refer to the threshold $\tau$ that achieves an FMR of $10^{-3}$ on the combined test set.} we reach TMR $=0.630$ for \texttt{original} and $0.897$ for \texttt{random}.
We can also observe the distribution change of impostor scores from \fig{rfwprotocol}, and \texttt{random} pushes impostor distributions of all ethnicities closer to 0 without changing the bias on \texttt{Caucasian}.
We suppose that impostor scores are not centered at 0 because of the performance limit of the network and the bias can be minimized if a good network is applied.

Cohort samples for RFW are taken from the BUPT-Balancedface \cite{wang2020mitigating} dataset, which has the same four ethnicities with 70000 subjects per ethnicity, but its image quality for \texttt{Asian} and \texttt{Indian} are not as good and stable as \texttt{African} and \texttt{Caucasian}, which have comparable quality to RFW images.
We cleaned the cohort dataset by removing possible overlaps with RFW and subjects that have different labels for duplicate images.
Since it is not precisely known how the impostor pairs of the default RFW protocol were selected, we rely on an IResNet100 network, which is trained on MS1M by ArcFace loss and knows RFW well,\footnote{\url{https://github.com/deepinsight/insightface}} to get 5000 most similar impostor pairs plus 5000 genuine pairs per ethnicity as the cohort for \texttt{original}.
The cohort for \texttt{random} is selected following the same idea as the test set.

%% file: texfiles/experiments.tex
\section{Experiments}
\label{sec:experiments}

\input{results/network.tex}
\begin{figure}[t]
  \centering
  \includegraphics[page=4,width=.45\textwidth,trim=0 0 0 10,clip]{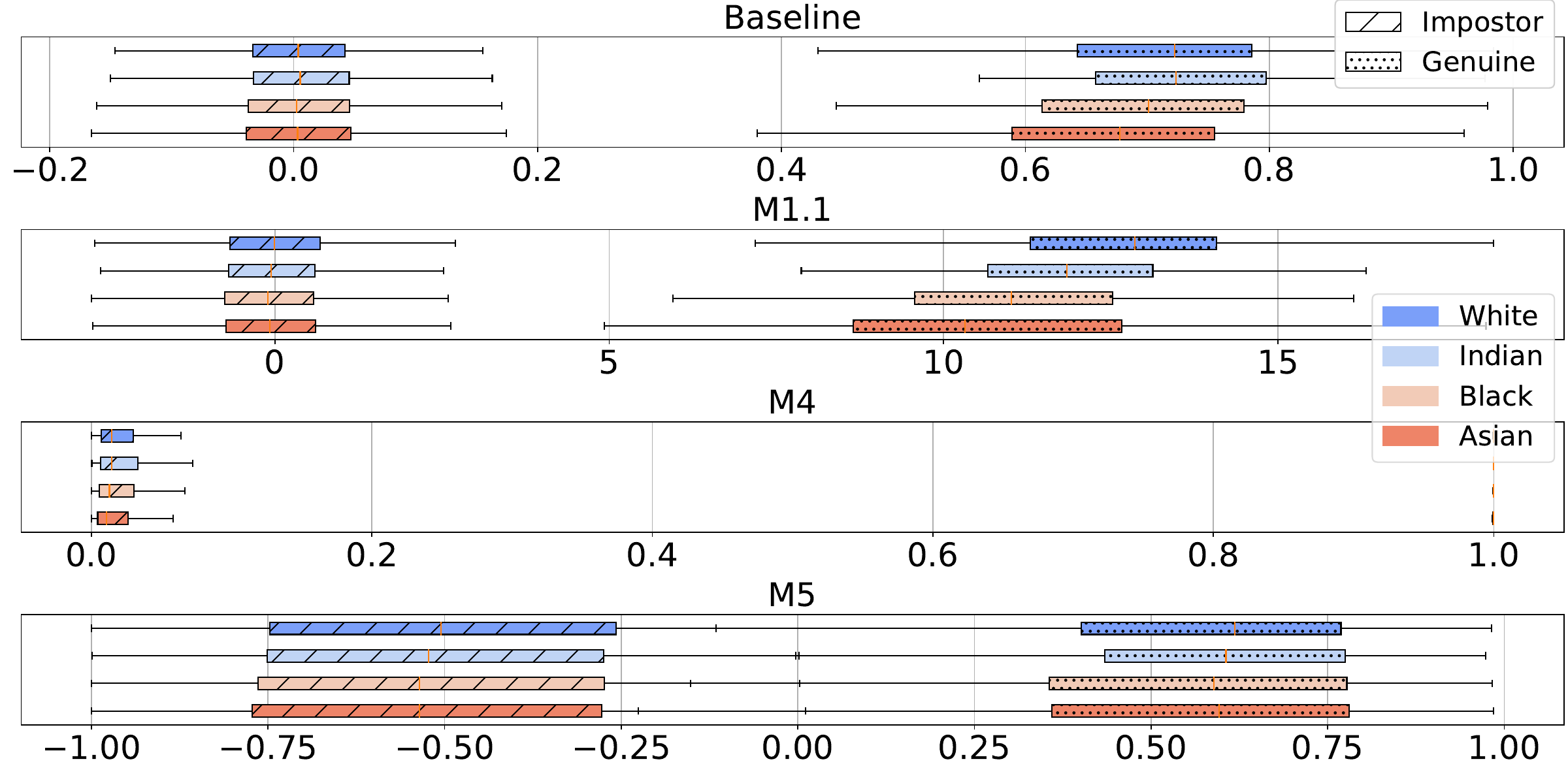}
  \Caption[fig:method_compare]{Impostor vs ALL}{
    This figure compares the VGG\-Face2 ethnicity score distributions of baseline, impostor-based method M1.1, and impostor-genuine-based methods M4 and M5. Features are extracted by E3.
}
\end{figure}

In total, we evaluate five different pre-trained and publicly available FR networks, as summarized in \tab{network}.
Since RFW is a subset of MS1M, any network that is trained on MS1M cannot be evaluated on RFW.
Therefore, we select two Arcface networks provided\footnote{\label{fn:rfwmodel}\url{http://www.whdeng.cn/RFW/model.html}} by the RFW authors \cite{wang2019racial}, \ie, a ResNet34 (E1) trained on CASIA-Webfaces, and a ResNet-50 architecture (E2) trained on MS1M excluding RFW identities.
Since other evaluated datasets do not have this issue, we also include a more powerful IResNet-100 topology trained on MS1M using MagFace loss (E4).\footnote{\url{https://github.com/IrvingMeng/MagFace}}
Finally, we employ two IResNet-100 architectures trained on the WebFaces12M dataset using AdaFace loss (E3)\footnote{\url{https://github.com/mk-minchul/AdaFace}} and DALIFace loss \cite{robbins2024daliid} (E5).
For comparison, we also evaluate FSN \cite{terhorst2020post} on our protocols and networks.

\subsection{VGGFace2}

We run tests on the VGGFace2 dataset with all feature extractors.
All proposed methods are applied, independently with two different types of demographic labels, gender and ethnicity.
In \fig{method_compare}, a quantile plot, the impostor distribution for baseline of VGGFace2 ethnicity is centered around 0, but a discrepancy appears in the genuine distribution.
Here the overlap of impostor and genuine scores are treated as outliers and discarded, while the overlap still exists.
The impostor-based method M1.1, which is supposed to only work on one side of the scores, results in some subtle moves on the impostor side, while genuine scores are stretched to more varied distributions.
FMR gaps between demographics are diminished, while FNMR gaps expand.
Theoretically, a small threshold should lead to a large FNMR, but WERM drops in \tab{compact} since the change in FMR dominates.
M4 and M5 attempt to harmonize the spread on both distributions simultaneously, and they perform differently on the task.
M4 provides nicely separated score distributions, but the alignment across demographics can be worse.
M5 models both distributions and achieves good alignment, but pushes the overlap of extreme values from both sides.

\input{results/znorm_compare_4num.tex}

\tab{compact} is a compact result table.
Each subtable displays TMR and WERM at $\tau=10^{-3}$ for scores post-processed by five demographic-based methods with feature extracted by E1 - E5.
An analysis for two factors of WERM is on our GitHub page.\footref{fn:github}
WERM for scores before normalization (baseline) are displayed in the first row of each table.
Almost all methods lead to a less biased output when normalized with respect to gender.
In most cases, verification performance is not affected or even has improvements, while the drop exists with a small magnitude.
The difficulty of alignment grows as the number of demographics rises to four ethnicities, especially M5 does not work well when facing three well-performing networks.
M1.1, M2.1, and M4 are quite stable and outperform FSN.
Comparison between pure identity- (M1) and demographic-based (M1.1/1.2) methods can be found in \tab{znorm}.
M1 does not exhibit a notable advantage over M1.1/1.2, which proves that demographic information is more influential in mitigating bias compared to identity-only data.
M1.1 and M1.2 have comparable performance, which is guaranteed when cohort size for M1.1 is large enough (Central Limit Theorem) for good estimation.

\input{results/compact_table_with_FSN.tex}

We observe that balancing ethnicity through sub-sampling worsens fairness compared to baselines, with a similar, albeit smaller, trend for gender balancing.
The unexpected rise after balancing can be attributed to the limited number of samples in the minority group.
Sub-sampling occurs in the majority groups and nearly all samples in the minority groups are preserved.
Thus, the distribution issue in the minority groups remains, the impact of the majority groups diminishes, and ultimately, the bias is amplified.
For the same reason, we observe a drop in TMR after balancing.
Similar behavior on both ethnicity and gender exhibits that balancing the impostor pairs per demographic via sub-sampling and then deciding thresholds does not lead to a less biased result.
Regardless of $\tau$ determination, our normalization techniques improve system fairness, though WERM magnitudes remain baseline-consistent. 
We prove that determining thresholds with balanced scores is unnecessary and may introduce extra bias.

\subsection{RFW}
\label{sec:exp:rfw}

All nine normalization techniques are applied to each protocol-network pair.
Although \texttt{random} is proposed to mitigate the selection bias in \texttt{original}, in \stab{rfw_o} and \subref{tab:rfw_r}, it is not the case by checking the WERM values for the baseline scores.
The bias seems not only brought by impostor score distribution and cannot be eliminated by the proposed new protocol.
To ensure the results for protocol \texttt{random} are not occasionally created, we perform a simplified statistical analysis by generating four more random splits and implement all proposed methods on them, and compute TMR and WERM values for all five splits.
For each network and method, we compute standard deviation (STD) across the five splits and compute the average across methods.
STD ranges from 0.04 to 2.748 for TMR and from 0.022 to 0.16 for WERM, which implies that no large variation for any method, so the results discussed below are reliable and reproducible.
The detailed table is available on the GitHub page.\footref{fn:github}

Across five methods, TMR undergoes changes ranging from $-0.8\%$ to $9.2\%$, while FSN lowers TMR in most cases.
Interestingly, normalization techniques exhibit distinct bias mitigation impacts on the two protocols.
E1, E2, and E5 yield decreasing WERM across all impostor-based methods (M1.1-M3), while M5 is preferable for \texttt{original}.
M4 only works well on features extracted by E1.
De-biasing effects brought by M1.1 are permanent for this dataset, with the enhanced TMR performance in most cases.
Consequently, the impostor-based methods, M1.1, M2.1, and M3 have steady achievement in de-biasing for both gender and ethnicity, regardless of datasets.
M4 is desired only for the VGGFace2 dataset, while M5 behaves positively for gender and the hardest RFW protocol.

%% file: results/network.tex
\begin{table}[t!]
    \footnotesize
    \centering
    \Caption[tab:network]{Pre-Trained Networks}{This table lists the networks utilized in our experiments, including data and loss function used for training. The networks are sorted in ascending order of overall recognition performance.}
    \resizebox{0.93\linewidth}{!}{%
    \begin{tabular}{c|c|c|c}\toprule
    \bf Model  &\bf Network  &\bf Training Data &\bf Loss Function \\\midrule
    E1 &ResNet34 &CASIA-WebFace &ArcFace  \\
    E2 &ResNet50 &MS1M-w/o-RFW &ArcFace  \\
    E3 &IResNet100 &Webface12M &AdaFace  \\
    E4 &IResNet100 &MS1M &MagFace  \\
    E5 &IResNet100 &Webface12M &DALIFace  \\
    \end{tabular}%
    }\vspace*{-2ex}
\end{table}

%% file: results/znorm_compare_4num.tex
\begin{table}[!t]
    \centering
    \Caption[tab:znorm]{Z-Norm-based Methods}{This table displays $\mathrm{WERM}$ and $\mathrm{TMR}$ (\%) of three Z-norm methods (M1, M1.1, M1.2) for VGGFace2 ethnicity w.r.t.~feature extractors E1 - E5. The best values per column are colored in blue/red.}
    \resizebox{\columnwidth}{!}{%
    \begin{tabular}{l|r@{\:}r|r@{\:}r|r@{\:}r|r@{\:}r|r@{\:}r}
    \toprule
    Network & \multicolumn{2}{c}{E1} & \multicolumn{2}{c}{E2} & \multicolumn{2}{c}{E3} & \multicolumn{2}{c}{E4} & \multicolumn{2}{c}{E5} \\\midrule
    Metrics & \small TMR $\uparrow$ & \small WERM $\downarrow$ & \small TMR $\uparrow$ & \small WERM $\downarrow$ & \small TMR $\uparrow$ & \small WERM $\downarrow$ & \small TMR $\uparrow$ & \small WERM $\downarrow$ & \small TMR $\uparrow$ & \small WERM $\downarrow$ \\
    \midrule
    Baseline & \textit{93.76} & \textit{3.8094} & \textit{95.65} & \textit{4.0101} & \textit{\textcolor{blue}{96.80}} & \textit{1.9320} & \textit{\textcolor{blue}{96.92}} & \textit{1.8356} & \textit{\textcolor{blue}{96.96}} & \textit{2.3542} \\
    M1 & \textcolor{blue}{94.50} & 1.6103 & \textcolor{blue}{95.81} & 3.7954 & 96.76 & 1.9046 & \textcolor{blue}{96.92} & \textcolor{red}{1.8154} & 96.88 & 3.1285 \\
    M1.1 & 92.85 & 1.5870 & 95.73 & \textcolor{red}{1.3594} & 96.76 & \textcolor{red}{1.6203} & 96.84 & 2.1640 & 96.88 & \textcolor{red}{1.6728} \\
    M1.2 & 92.98 & \textcolor{red}{1.5237} & 95.56 & 2.5367 & \textcolor{blue}{96.80} & 1.7845 & 96.88 & 1.9265 & 96.80 & 2.6092
\end{tabular}%
    }\vspace*{-4ex}
\end{table}

%% file: results/compact_table_with_FSN.tex
\begin{table*}[p]
  \centering
  \Caption[tab:compact]{$\mathrm{WERM_{10^{-3}}}$ \& $\mathrm{TMR_{10^{-3}}}$}{
    Six tables below present results for six evaluation protocols.
    For each protocol, all five networks, E1 - E5 as provided in \tab{network}, are used to extract features.
    Five proposed score normalization methods M1 -- M5, cf.~\tab{method}, and FSN \cite{terhorst2020post}, are applied to those features and TMR (\%) and WERM values at threshold $\tau=10^{-3}$ are computed.
    \textcolor{blue}{best} and \textcolor{cyan}{runner-up} TMR and \textcolor{red}{best} and \textcolor{orange}{second} WERM value are highlighted.
    The \textit{first row} shows the baseline.
  }\vspace*{1ex}

  \subfloat[VGGFace2 Gender\label{tab:vgg_g}]{
    \centering
    \resizebox{0.72\textwidth}{!}{%
    \begin{tabular}{l|rr|rr|rr|rr|rrr}
    \toprule
    Network & \multicolumn{2}{c}{E1} & \multicolumn{2}{c}{E2} & \multicolumn{2}{c}{E3} & \multicolumn{2}{c}{E4} & \multicolumn{2}{c}{E5} \\\midrule
    Metrics & TMR $\uparrow$ & WERM $\downarrow$ & TMR $\uparrow$ & WERM $\downarrow$ & TMR $\uparrow$ & WERM $\downarrow$ & TMR $\uparrow$ & WERM $\downarrow$ & TMR $\uparrow$ & WERM $\downarrow$ \\
    \midrule
    Baseline & \textit{\textcolor{cyan}{93.55}} & \textit{1.6477} & \textit{95.48} & \textit{1.4042} & \textit{\textcolor{cyan}{96.71}} & \textit{1.2908} & \textit{\textcolor{cyan}{96.88}} & \textit{1.1996} & \textit{96.76} & \textit{1.3892} \\
    FSN & 92.32 & 1.7575 & 95.36 & 1.3947 & \textcolor{blue}{96.76} & 1.3059 & 96.71 & 1.1674 & 96.84 & 1.4596 \\
    M1.1 & 93.43 & \textcolor{red}{1.1649} & \textcolor{cyan}{95.65} & \textcolor{orange}{1.0977} & 96.67 & 1.0997 & \textcolor{cyan}{96.88} & 1.0986 & \textcolor{blue}{96.92} & \textcolor{orange}{1.0831} \\
    M2.1 & \textcolor{blue}{93.63} & 1.2033 & \textcolor{blue}{95.77} & \textcolor{red}{1.0932} & 96.63 & 1.1092 & \textcolor{blue}{96.92} & \textcolor{red}{1.0349} & 96.84 & 1.0926 \\
    M3 & 92.98 & \textcolor{orange}{1.1779} & 95.32 & 1.1346 & 96.67 & \textcolor{red}{1.0445} & 96.84 & \textcolor{orange}{1.0366} & 96.80 & \textcolor{red}{1.0476} \\
    M4 & 93.35 & 1.3064 & 95.36 & 1.1233 & \textcolor{cyan}{96.71} & 1.1540 & \textcolor{cyan}{96.88} & 1.1420 & \textcolor{cyan}{96.88} & 1.1289 \\
    M5 & \textcolor{cyan}{93.55} & 1.3906 & 95.44 & 1.1811 & 96.67 & \textcolor{orange}{1.0505} & \textcolor{cyan}{96.88} & 1.0474 & \textcolor{cyan}{96.88} & 1.1871 \\
 
    \end{tabular}%
    }
  }\vspace*{-1ex}

  \subfloat[VGGFace2 Gender Balanced\label{tab:vgg_gb}]{
    \centering
    \resizebox{0.72\textwidth}{!}{%
    \begin{tabular}{l|rr|rr|rr|rr|rrr}
    \toprule
    Network & \multicolumn{2}{c}{E1} & \multicolumn{2}{c}{E2} & \multicolumn{2}{c}{E3} & \multicolumn{2}{c}{E4} & \multicolumn{2}{c}{E5} \\\midrule
    Metrics & TMR $\uparrow$ & WERM $\downarrow$ & TMR $\uparrow$ & WERM $\downarrow$ & TMR $\uparrow$ & WERM $\downarrow$ & TMR $\uparrow$ & WERM $\downarrow$ & TMR $\uparrow$ & WERM $\downarrow$ \\
    \midrule
    Baseline & \textit{93.02} & \textit{1.7456} & \textit{95.40} & \textit{1.4696} & \textit{\textcolor{cyan}{96.63}} & \textit{1.2624} & \textit{96.80} & \textit{1.2490} & \textit{96.71} & \textit{1.3916} \\
    FSN & 91.46 & 1.7486 & 95.11 & 1.4429 & \textcolor{blue}{96.67} & 1.2761 & 96.71 & 1.1664 & 96.59 & 1.4267 \\
    M1.1 & \textcolor{cyan}{93.35} & \textcolor{red}{1.1637} & \textcolor{cyan}{95.65} & \textcolor{orange}{1.1080} & \textcolor{blue}{96.67} & \textcolor{orange}{1.0525} & 96.76 & 1.0811 & \textcolor{blue}{96.92} & \textcolor{orange}{1.1017} \\
    M2.1 & \textcolor{blue}{93.55} & 1.2197 & \textcolor{blue}{95.77} & \textcolor{red}{1.1014} & \textcolor{cyan}{96.63} & 1.0686 & \textcolor{blue}{96.92} & \textcolor{red}{1.0425} & 96.84 & 1.1097 \\
    M3 & 92.98 & \textcolor{orange}{1.1723} & 95.32 & 1.1426 & \textcolor{blue}{96.67} & \textcolor{red}{1.0408} & 96.84 & \textcolor{orange}{1.0455} & 96.80 & \textcolor{red}{1.0688} \\
    M4 & 93.14 & 1.3243 & 95.32 & 1.1330 & \textcolor{cyan}{96.63} & 1.1258 & \textcolor{cyan}{96.88} & 1.1642 & 96.80 & 1.1633 \\
    M5 & 93.06 & 1.4569 & 95.28 & 1.2022 & \textcolor{blue}{96.67} & \textcolor{red}{1.0408} & \textcolor{cyan}{96.88} & 1.0483 & \textcolor{cyan}{96.88} & 1.2092 \\

    \end{tabular}%
    }
  }\vspace*{-1ex}

  \subfloat[VGGFace2 Ethnicity\label{tab:vgg_r}]{
    \centering
    \resizebox{0.72\textwidth}{!}{%
    \begin{tabular}{l|rr|rr|rr|rr|rrr}
    \toprule
    Network & \multicolumn{2}{c}{E1} & \multicolumn{2}{c}{E2} & \multicolumn{2}{c}{E3} & \multicolumn{2}{c}{E4} & \multicolumn{2}{c}{E5} \\\midrule
    Metrics & TMR $\uparrow$ & WERM $\downarrow$ & TMR $\uparrow$ & WERM $\downarrow$ & TMR $\uparrow$ & WERM $\downarrow$ & TMR $\uparrow$ & WERM $\downarrow$ & TMR $\uparrow$ & WERM $\downarrow$ \\
    \midrule
    Baseline & \textit{93.76} & \textit{3.8094} & \textit{95.65} & \textit{4.0101} & \textit{\textcolor{cyan}{96.80}} & \textit{1.9320} & \textit{\textcolor{cyan}{96.92}} & \textit{1.8356} & \textit{\textcolor{blue}{96.96}} & \textit{2.3542} \\
    FSN & 92.69 & 3.3962 & 95.65 & 4.0190 & 96.63 & 1.9067 & \textcolor{cyan}{96.92} & \textcolor{orange}{1.7605} & 96.84 & 4.9612 \\
    M1.1 & 92.85 & \textcolor{red}{1.5870} & \textcolor{blue}{95.73} & \textcolor{red}{1.3594} & 96.76 & \textcolor{orange}{1.6203} & 96.84 & 2.1640 & 96.88 & \textcolor{red}{1.6728} \\
    M2.1 & 92.28 & \textcolor{orange}{2.1288} & \textcolor{cyan}{95.69} & \textcolor{orange}{1.4253} & 96.71 & \textcolor{red}{1.3468} & \textcolor{blue}{96.96} & 2.9963 & 96.80 & 2.0942 \\
    M3 & 92.98 & 2.5158 & 95.52 & 2.6485 & \textcolor{cyan}{96.80} & 1.6644 & 96.88 & 1.8598 & 96.80 & 2.4600 \\
    M4 & \textcolor{blue}{94.00} & 2.6577 & \textcolor{cyan}{95.69} & 3.7550 & \textcolor{blue}{96.84} & 1.7912 & \textcolor{cyan}{96.92} & \textcolor{red}{1.6990} & \textcolor{blue}{96.96} & \textcolor{orange}{1.9823} \\
    M5 & \textcolor{cyan}{93.92} & 3.4323 & 95.65 & 3.6006 & \textcolor{cyan}{96.80} & 1.9328 & 96.88 & 1.9968 & \textcolor{cyan}{96.92} & 3.9164 \\

    \end{tabular}%
    }
  }\vspace*{-1ex}

  \subfloat[VGGFace2 Ethnicity Balanced\label{tab:vgg_rb}]{
    \centering
    \resizebox{0.72\textwidth}{!}{%
    \begin{tabular}{l|rr|rr|rr|rr|rrr}
    \toprule
    Network & \multicolumn{2}{c}{E1} & \multicolumn{2}{c}{E2} & \multicolumn{2}{c}{E3} & \multicolumn{2}{c}{E4} & \multicolumn{2}{c}{E5} \\\midrule
    Metrics & TMR $\uparrow$ & WERM $\downarrow$ & TMR $\uparrow$ & WERM $\downarrow$ & TMR $\uparrow$ & WERM $\downarrow$ & TMR $\uparrow$ & WERM $\downarrow$ & TMR $\uparrow$ & WERM $\downarrow$ \\
    \midrule
    Baseline & \textit{88.62} & \textit{6.4436} & \textit{94.25} & \textit{6.0192} & \textit{96.51} & \textit{2.9195} & \textit{96.47} & \textit{2.7463} & \textit{96.39} & \textit{3.2102} \\
    FSN & 86.57 & 11.1599 & 94.05 & 5.2970 & 96.35 & 3.8272 & 96.38 & \textcolor{orange}{2.0060} & 96.35 & 3.9992 \\
    M1.1 & \textcolor{cyan}{93.26} & \textcolor{red}{2.0316} & \textcolor{blue}{95.85} & \textcolor{orange}{2.0109} & \textcolor{cyan}{96.71} & \textcolor{orange}{1.9065} & \textcolor{cyan}{96.88} & 2.5114 & \textcolor{blue}{96.88} & \textcolor{red}{1.9237} \\
    M2.1 & 93.18 & \textcolor{orange}{2.5456} & \textcolor{blue}{95.85} & \textcolor{red}{1.8519} & \textcolor{blue}{96.76} & \textcolor{red}{1.7582} & \textcolor{blue}{96.96} & 2.5224 & \textcolor{blue}{96.88} & \textcolor{orange}{2.4882} \\
    M3 & \textcolor{blue}{93.43} & 3.3406 & \textcolor{cyan}{95.69} & 2.6263 & \textcolor{blue}{96.76} & 2.3642 & 96.76 & 2.9416 & \textcolor{cyan}{96.80} & 2.5019 \\
    M4 & 92.65 & 3.6487 & 95.03 & 2.8245 & 96.51 & 1.9404 & 96.59 & \textcolor{red}{1.6599} & 96.67 & 3.0813 \\
    M5 & 88.91 & 6.1117 & 93.72 & 6.0567 & 96.22 & 3.6257 & 95.81 & 3.3750 & 95.73 & 5.6425 \\

    \end{tabular}%
    }
  }\vspace*{-1ex}

  \subfloat[RFW Original\label{tab:rfw_o}]{
    \centering
    \resizebox{0.72\textwidth}{!}{%
    \begin{tabular}{l|rr|rr|rr|rr|rrr}
    \toprule
    Network & \multicolumn{2}{c}{E1} & \multicolumn{2}{c}{E2} & \multicolumn{2}{c}{E3} & \multicolumn{2}{c}{E4} & \multicolumn{2}{c}{E5} \\\midrule
    Metrics & TMR $\uparrow$ & WERM $\downarrow$ & TMR $\uparrow$ & WERM $\downarrow$ & TMR $\uparrow$ & WERM $\downarrow$ & TMR $\uparrow$ & WERM $\downarrow$ & TMR $\uparrow$ & WERM $\downarrow$ \\
    \midrule
    Baseline & \textit{24.22} & \textit{2.5246} & \textit{63.05} & \textit{2.5402} & \textit{89.14} & \textit{3.4611} & --- & --- & \textit{89.03} & \textit{2.7167} \\
    FSN & 2.19 & 6.8737 & 55.81 & 3.0454 & 88.60 & 4.5858 & --- & --- & 87.85 & 3.6184 \\
    M1.1 & \textcolor{cyan}{33.11} & \textcolor{orange}{1.6128} & \textcolor{blue}{67.13} & \textcolor{orange}{2.0791} & 89.20 & \textcolor{red}{2.7097} & --- & --- & 88.59 & 2.4684 \\
    M2.1 & \textcolor{blue}{33.38} & \textcolor{red}{1.4072} & \textcolor{cyan}{65.35} & \textcolor{red}{2.0301} & \textcolor{blue}{90.41} & 7.0568 & --- & --- & 89.54 & 2.2140 \\
    M3 & 27.99 & 1.7419 & 62.17 & 2.2668 & 88.92 & 3.8758 & --- & --- & \textcolor{cyan}{89.56} & 2.1883 \\
    M4 & 26.29 & 2.0549 & 62.25 & 2.7727 & 89.60 & 3.5602 & --- & --- & \textcolor{blue}{89.91} & \textcolor{orange}{1.8976} \\
    M5 & 25.63 & 2.2587 & 62.54 & 2.5204 & \textcolor{cyan}{90.05} & \textcolor{orange}{3.1712} & --- & --- & 89.50 & \textcolor{red}{1.6877} \\

    \end{tabular}%
    }
  }\vspace*{-1ex}

  \subfloat[RFW Random\label{tab:rfw_r}]{
    \centering
    \resizebox{0.72\textwidth}{!}{%
    \begin{tabular}{l|rr|rr|rr|rr|rrr}
    \toprule
    Network & \multicolumn{2}{c}{E1} & \multicolumn{2}{c}{E2} & \multicolumn{2}{c}{E3} & \multicolumn{2}{c}{E4} & \multicolumn{2}{c}{E5} \\\midrule
    Metrics & TMR $\uparrow$ & WERM $\downarrow$ & TMR $\uparrow$ & WERM $\downarrow$ & TMR $\uparrow$ & WERM $\downarrow$ & TMR $\uparrow$ & WERM $\downarrow$ & TMR $\uparrow$ & WERM $\downarrow$ \\
    \midrule
    Baseline & \textit{60.25} & \textit{2.0418} & \textit{89.66} & \textit{2.7152} & \textit{98.08} & \textit{2.3202} & --- & --- & \textit{98.04} & \textit{4.3784} \\
    FSN & 56.15 & 3.4326 & 87.89 & 3.2014 & \textcolor{cyan}{98.10} & 3.0989 & --- & --- & 98.17 & 7.6021 \\
    M1.1 & \textcolor{blue}{68.35} & 1.9580 & \textcolor{cyan}{90.55} & \textcolor{red}{1.7059} & \textcolor{blue}{98.37} & \textcolor{red}{1.5468} & --- & --- & \textcolor{blue}{98.66} & 3.2601 \\
    M2.1 & 63.53 & \textcolor{red}{1.4274} & 90.31 & \textcolor{orange}{1.9231} & 97.97 & 1.7707 & --- & --- & \textcolor{cyan}{98.59} & \textcolor{orange}{3.1609} \\
    M3 & \textcolor{cyan}{64.25} & 1.7578 & 89.30 & 2.2723 & 98.09 & \textcolor{orange}{1.6354} & --- & --- & 98.42 & \textcolor{red}{2.0688} \\
    M4 & 64.11 & \textcolor{orange}{1.7085} & \textcolor{blue}{90.84} & 2.6720 & 98.06 & 2.5022 & --- & --- & 98.31 & 4.8942 \\
    M5 & 60.57 & 2.5452 & 86.83 & 3.4545 & 97.46 & 6.7419 & --- & --- & 97.20 & 5.0545 \\

    \end{tabular}%
    }
  }
 \end{table*}

%% file: texfiles/discussion.tex
\section{Discussion}
\label{sec:discussion}

In WERM \eqref{eq:werm}, $\alpha=\frac12$ is set to balance the contribution of FNMR and FMR, but balancing is not guaranteed by $\alpha$ alone.
We compute the relative contribution of FMR and FNMR ($R_{\mathrm{FMR}}$, $R_{\mathrm{FNMR}}$, respectively) in \eqref{eq:werm} by dividing scaled worst-case FMR and FNMR by WERM, take $\delta=R_{\mathrm{FMR}}-R_{\mathrm{FNMR}}$, and then analyze the distribution of $\delta$ with respect to each method.
A higher $\delta$ implies a lower $R_{\mathrm{FNMR}}$.
Our hypothesis is confirmed by the distribution plot in \fig{contr_diff}, where genuine-impostor-based methods (M4, M5) and identity-based methods (M1, M2) more frequently result in a large $\delta$ than the other methods.
Most methods have $\delta$ centered lower than baselines.
Protocols like VGGFace2 gender-balanced with all decreases on WERM locates mostly at the lower tail, and method like M5 has a large portion of $\delta$ clustered above 0.5 which is consistent with \tab{compact}.
Impostor-based methods (M1-M3), which focus on aligning impostor scores, lead to a smaller FMR and higher FNMR differences so that $\delta$ decreases.
However, for genuine-impostor-based methods, taking care of both sides simultaneously does not ensure the change in two error rates will be the same.
Some unpredictable effects occur depending on the dominant side.
For example, better alignment in the genuine side results in higher FMR and smaller FNMR differences, leading to a larger $\delta$, or vice versa.
In general, impostor-based methods are more stable in de-biasing at scoring time.

\begin{figure}[t]
  \centering
  \includegraphics[page=1,width=.45\textwidth,clip]{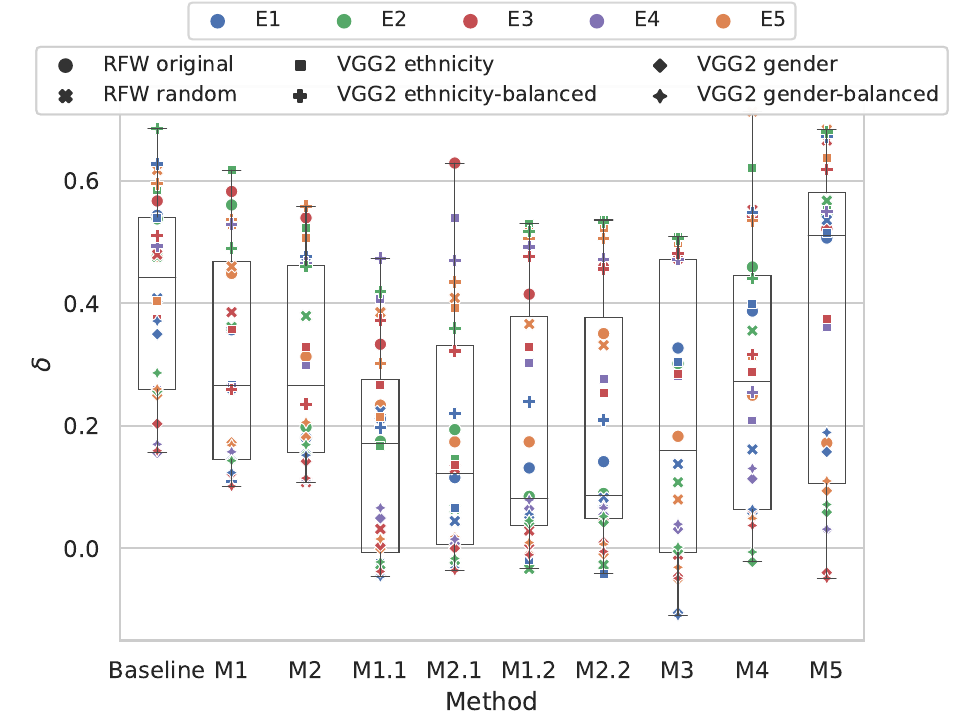}
  \Caption[fig:contr_diff]{Distribution of $\delta$}{
    This figure exhibits the distribution of FMR and FNMR contribution difference $\delta$ with respect to the baseline and each method.
}
\end{figure}

%% file: texfiles/conclusion.tex
\section{Conclusion}
\label{sec:conclusion}

We propose nine score normalization techniques, two are well-known identity-based methods (M1, M2), each followed by two extensions that integrate demographic information (M1.1, M1.2, M2.1, M2.2), and three pure cohort-based methods (M3, M4, M5).
All techniques improve demographic fairness for high-security applications, \ie, at low FMR, by working solely on scores without requiring network or feature adaptation.
Importantly, in opposition to many feature-based fairness improvement techniques, none of our methods decreases verification performance, even small improvements can be observed.
Experiments on six protocols from two datasets and five pre-trained feature extractors demonstrate the consistency of impostor-based methods (M1.1, M1.2, M3) with different verification performances.
Analysis of the WERM value reveals the unequal contribution of FNMR and FMR in fairness evaluation, which is planned to be improved next.
Also, other distributions than Normal will be explored to capture tail behavior.

%% file: main.bbl
\begin{thebibliography}{10}\itemsep=-1pt

\bibitem{albiero2020sexist}
V.~Albiero and K.~W. Bowyer.
\newblock Is face recognition sexist? no, gendered hairstyles and biology are.
\newblock In {\em British Machine Vision Virtual Conference {BMVC}}. {BMVA}
  Press, 2020.

\bibitem{albiero2020gender}
V.~Albiero, K.~KS, K.~Vangara, K.~Zhang, M.~C. King, and K.~W. Bowyer.
\newblock Analysis of gender inequality in face recognition accuracy.
\newblock In {\em Winter Conference on Applications of Computer Vision (WACV)
  Workshops}, pages 81--89. IEEE/CVF, 2020.

\bibitem{albiero2020balance}
V.~Albiero, K.~Zhang, and K.~W. Bowyer.
\newblock How does gender balance in training data affect face recognition
  accuracy?
\newblock In {\em International Joint Conference on Biometrics (IJCB)}, 2020.

\bibitem{atzori2023demographic}
A.~Atzori, G.~Fenu, and M.~Marras.
\newblock Demographic bias in low-resolution deep face recognition in the wild.
\newblock {\em IEEE Journal of Selected Topics in Signal Processing}, 2023.

\bibitem{auckenthaler2004score}
R.~Auckenthaler, M.~Carey, and H.~Lloyd-Thomas.
\newblock Score normalization for text-independent speaker verification
  systems.
\newblock {\em Digital Signal Processing}, 10(1), 2000.

\bibitem{cao2018vggface2}
Q.~Cao, L.~Shen, W.~Xie, O.~M. Parkhi, and A.~Zisserman.
\newblock {VGGFace2}: A dataset for recognising faces across pose and age.
\newblock In {\em Automatic Face {\&} Gesture Recognition (FG)}. IEEE, 2018.

\bibitem{cavazos2020accuracy}
J.~G. Cavazos, P.~J. Phillips, C.~D. Castillo, and A.~J. O’Toole.
\newblock Accuracy comparison across face recognition algorithms: Where are we
  on measuring race bias?
\newblock {\em Transactions on Biometrics, Behavior, and Identity Science
  (TBIOM)}, 3(1):101--111, 2020.

\bibitem{cook2019demographic}
C.~M. Cook, J.~J. Howard, Y.~B. Sirotin, J.~L. Tipton, and A.~R. Vemury.
\newblock Demographic effects in facial recognition and their dependence on
  image acquisition: An evaluation of eleven commercial systems.
\newblock {\em Transactions on Biometrics, Behavior, and Identity Science
  (TBIOM)}, 1(1):32--41, 2019.

\bibitem{pereira2021fairness}
T.~de~Freitas~Pereira and S.~Marcel.
\newblock Fairness in biometrics: A figure of merit to assess biometric
  verification systems.
\newblock {\em Transactions on Biometrics, Behavior, and Identity Science
  (TBIOM)}, 4(1):19--29, 2021.

\bibitem{delrio2016automated}
J.~S. del Rio, D.~Moctezuma, C.~Conde, I.~M. de~Diego, and E.~Cabello.
\newblock Automated border control e-gates and facial recognition systems.
\newblock {\em Computers \& Security}, 62, 2016.

\bibitem{deng2019arcface}
J.~Deng, J.~Guo, N.~Xue, and S.~Zafeiriou.
\newblock Arcface: Additive angular margin loss for deep face recognition.
\newblock In {\em Conference on Computer Vision and Pattern Recognition
  (CVPR)}, 2019.

\bibitem{doddington1998sheep}
G.~R. Doddington, W.~Liggett, A.~F. Martin, M.~A. Przybocki, and D.~A.
  Reynolds.
\newblock {SHEEP, GOATS, LAMBS and WOLVES}: A statistical analysis of speaker
  performance in the {NIST} 1998 speaker recognition evaluation.
\newblock In {\em International Conference on Spoken Language Processing
  (ICSPL)}, 1998.

\bibitem{sukthanker2022importance}
S.~Dooley, R.~S. Sukthanker, J.~P. Dickerson, C.~White, F.~Hutter, and
  M.~Goldblum.
\newblock On the importance of architectures and hyperparameters for fairness
  in face recognition.
\newblock In {\em Workshop on Trustworthy and Socially Responsible Machine
  Learning, NeurIPS 2022}, 2022.

\bibitem{duta2021iresnet}
I.~C. Duta, L.~Liu, F.~Zhu, and L.~Shao.
\newblock Improved residual networks for image and video recognition.
\newblock In {\em International Conference on Pattern Recognition (ICPR)},
  pages 9415--9422. IEEE, 2021.

\bibitem{gong2020jointly}
S.~Gong, X.~Liu, and A.~K. Jain.
\newblock Jointly de-biasing face recognition and demographic attribute
  estimation.
\newblock In {\em European Conference on Computer Vision (ECCV)}. Springer,
  2020.

\bibitem{gong2021mitigating}
S.~Gong, X.~Liu, and A.~K. Jain.
\newblock Mitigating face recognition bias via group adaptive classifier.
\newblock In {\em Conference on Computer Vision and Pattern Recognition
  (CVPR)}, 2021.

\bibitem{grother2021demographic}
P.~Grother.
\newblock Demographic differentials in face recognition algorithms.
\newblock {\em Virtual Events Series--Demo-Graphic Fairness in Biometric
  Systems}, 2021.

\bibitem{grother2022nist}
P.~Grother.
\newblock Face recognition vendor test {(FRVT)} part 8: Summarizing demographic
  differentials.
\newblock Technical report, National Institute of Standards and Technology
  (NIST), 2022.

\bibitem{grother2019frvt}
P.~Grother, M.~Ngan, and K.~Hanaoka.
\newblock Face recognition vendor test ({FRVT}) part 3: Demographic effects.
\newblock Technical report, National Institute of Standards and Technology
  (NIST), 2018.

\bibitem{guenther2009maximumlikelihood}
M.~G\"unther and R.~P. W{\"u}rtz.
\newblock Face detection and recognition using maximum likelihood classifiers
  on {Gabor} graphs.
\newblock {\em International Journal of Pattern Recognition and Artificial
  Intelligence (IJPRAI)}, 23(03):433--461, 2009.

\bibitem{guo2016msceleb1m}
Y.~Guo, L.~Zhang, Y.~Hu, X.~He, and J.~Gao.
\newblock {MS-Celeb-1M}: A dataset and benchmark for large-scale face
  recognition.
\newblock In {\em European Conference on Computer Vision (ECCV)}. Springer,
  2016.

\bibitem{he2016deep}
K.~He, X.~Zhang, S.~Ren, and J.~Sun.
\newblock Deep residual learning for image recognition.
\newblock In {\em Conference on Computer Vision and Pattern Recognition
  (CVPR)}. IEEE, 2016.

\bibitem{hill2020wrongfully}
K.~Hill.
\newblock Wrongfully accused by an algorithm.
\newblock {\em New York Times}, 6 2020.
\newblock
  \url{https://www.nytimes.com/2020/06/24/technology/facial-recognition-arrest.html}.

\bibitem{howard2022evaluating}
J.~J. Howard, E.~J. Laird, R.~E. Rubin, Y.~B. Sirotin, J.~L. Tipton, and A.~R.
  Vemury.
\newblock Evaluating proposed fairness models for face recognition algorithms.
\newblock In {\em International Conference on Pattern Recognition (ICPR)},
  pages 431--447. Springer, 2022.

\bibitem{howard2019effect}
J.~J. Howard, Y.~B. Sirotin, and A.~R. Vemury.
\newblock The effect of broad and specific demographic homogeneity on the
  imposter distributions and false match rates in face recognition algorithm
  performance.
\newblock In {\em International Conference on Biometrics Theory, Applications
  and Systems (BTAS)}. IEEE, 2019.

\bibitem{hu2018squeeze}
J.~Hu, L.~Shen, and G.~Sun.
\newblock Squeeze-and-excitation networks.
\newblock In {\em Conference on Computer Vision and Pattern Recognition
  (CVPR)}, 2018.

\bibitem{huang2019deep}
C.~Huang, Y.~Li, C.~C. Loy, and X.~Tang.
\newblock Deep imbalanced learning for face recognition and attribute
  prediction.
\newblock {\em Transactions on Pattern Analysis and Machine Intelligence
  (TPAMI)}, 2019.

\bibitem{kim2022adaface}
M.~Kim, A.~K. Jain, and X.~Liu.
\newblock {AdaFace}: Quality adaptive margin for face recognition.
\newblock In {\em Conference on Computer Vision and Pattern Recognition
  (CVPR)}, 2022.

\bibitem{kortylewski2019analyzing}
A.~Kortylewski, B.~Egger, A.~Schneider, T.~Gerig, A.~Morel-Forster, and
  T.~Vetter.
\newblock Analyzing and reducing the damage of dataset bias to face recognition
  with synthetic data.
\newblock In {\em Conference on Computer Vision and Pattern Recognition
  (CVPR)}, 2019.

\bibitem{kotwal2022fairness}
K.~Kotwal and S.~Marcel.
\newblock Fairness index measures to evaluate bias in biometric recognition.
\newblock In {\em International Conference on Pattern Recognition Workshops
  (ICPRW)}, 2022.

\bibitem{kotwal2024mitigating}
K.~Kotwal and S.~Marcel.
\newblock Mitigating demographic bias in face recognition via regularized score
  calibration.
\newblock In {\em Winter Conference on Applications of Computer Vision (WACV)},
  2024.

\bibitem{krishnapriya2020skintone}
K.~S. Krishnapriya, V.~Albiero, K.~Vangara, M.~C. King, and K.~W. Bowyer.
\newblock Issues related to face recognition accuracy varying based on race and
  skin tone.
\newblock {\em Transactions on Technology and Society (TTS)}, 1(1):8--20, 2020.

\bibitem{liang2019additive}
J.~Liang, Y.~Cao, C.~Zhang, S.~Chang, K.~Bai, and Z.~Xu.
\newblock Additive adversarial learning for unbiased authentication.
\newblock In {\em Conference on Computer Vision and Pattern Recognition
  (CVPR)}, 2019.

\bibitem{liu2017sphereface}
W.~Liu, Y.~Wen, Z.~Yu, M.~Li, B.~Raj, and L.~Song.
\newblock {SphereFace}: Deep hypersphere embedding for face recognition.
\newblock In {\em Conference on Computer Vision and Pattern Recognition
  (CVPR)}, 2017.

\bibitem{malpass1969recognition}
R.~S. Malpass and J.~Kravitz.
\newblock Recognition for faces of own and other race.
\newblock {\em Journal of Personality and Social Psychology}, 1969.

\bibitem{mandasari2014calibration}
M.~I. Mandasari, M.~G\"unther, R.~Wallace, R.~Saeidi, S.~Marcel, and D.~A. van
  Leeuwen.
\newblock Score calibration in face recognition.
\newblock {\em IET Biometrics}, 3(4):246--256, 2014.

\bibitem{maze2018ijbc}
B.~Maze, J.~Adams, J.~A. Duncan, N.~Kalka, T.~Miller, C.~Otto, A.~K. Jain,
  W.~T. Niggel, J.~Anderson, J.~Cheney, and P.~Grother.
\newblock {IARPA} {Janus} {Benchmark - C}: Face dataset and protocol.
\newblock In {\em International Conference on Biometrics (ICB)}, 2018.

\bibitem{meng2021magface}
Q.~Meng, S.~Zhao, Z.~Huang, and F.~Zhou.
\newblock {MagFace}: A universal representation for face recognition and
  quality assessment.
\newblock In {\em Conference on Computer Vision and Pattern Recognition
  (CVPR)}, 2021.

\bibitem{moghaddam1998beyond}
B.~Moghaddam, W.~Wahid, and A.~Pentland.
\newblock Beyond eigenfaces: Probabilistic matching for face recognition.
\newblock In {\em International Conference on Automatic Face and Gesture
  Recognition (FG)}. IEEE, 1998.

\bibitem{morales2020sensitivenets}
A.~Morales, J.~Fierrez, R.~Vera-Rodriguez, and R.~Tolosana.
\newblock {SensitiveNets}: Learning agnostic representations with application
  to face images.
\newblock {\em Transactions on Pattern Analysis and Machine Intelligence
  (TPAMI)}, 2020.

\bibitem{phillips2011evaluation}
P.~J. Phillips, P.~Grother, and R.~Micheals.
\newblock {\em Handbook of Face Recognition}, chapter Evaluation Methods in
  Face Recognition.
\newblock Springer, 2nd edition, 2011.

\bibitem{platt1999probabilistic}
J.~Platt.
\newblock Probabilistic outputs for support vector machines and comparisons to
  regularized likelihood methods.
\newblock {\em Advances in Large Margin Classifiers}, 10(3):61--74, 1999.

\bibitem{poh2005f}
N.~Poh and S.~Bengio.
\newblock F-ratio client dependent normalisation for biometric authentication
  tasks.
\newblock In {\em International Conference on Acoustics, Speech, and Signal
  Processing (ICASSP)}. IEEE, 2005.

\bibitem{poh2010group}
N.~Poh, J.~Kittler, A.~Rattani, and M.~Tistarelli.
\newblock Group-specific score normalization for biometric systems.
\newblock In {\em Conference on Computer Vision and Pattern Recognition (CVPR)
  Workshops}, pages 38--45. IEEE, 2010.

\bibitem{poh2009adaptive}
N.~Poh, A.~Merati, and J.~Kittler.
\newblock Adaptive client-impostor centric score normalization: A case study in
  fingerprint verification.
\newblock In {\em International Conference on Biometrics: Theory, Applications,
  and Systems (BTAS)}. IEEE, 2009.

\bibitem{qin2020asymmetric}
H.~Qin.
\newblock Asymmetric rejection loss for fairer face recognition.
\newblock {\em arXiv}, 2020.

\bibitem{rahimi2023toward}
P.~Rahimi, C.~Ecabert, and S.~Marce.
\newblock Toward responsible face datasets: modeling the distribution of a
  disentangled latent space for sampling face images from demographic groups.
\newblock In {\em 2023 IEEE International Joint Conference on Biometrics
  (IJCB)}, pages 1--11. IEEE, 2023.

\bibitem{reynolds2000speaker}
D.~A. Reynolds, T.~F. Quatieri, and R.~B. Dunn.
\newblock Speaker verification using adapted gaussian mixture models.
\newblock {\em Digital Signal Processing}, 10(1):19--41, 2000.

\bibitem{robbins2024daliid}
W.~Robbins, G.~Bertocco, and T.~E. Boult.
\newblock {DaliID}: Distortion-adaptive learned invariance for identification
  – a robust technique for face recognition and person re-identification.
\newblock {\em IEEE Access}, 2024.

\bibitem{romm2018amazons}
T.~Romm.
\newblock Amazon’s facial-recognition tool misidentified 28 lawmakers as
  people arrested for a crime, study finds.
\newblock {\em Washington Post}, July 2018.
\newblock Retrieved from \url{http://www.washingtonpost.com}.

\bibitem{rudd2016moon}
E.~M. Rudd, M.~G\"unther, and T.~E. Boult.
\newblock {MOON}: A mixed objective optimization network for the recognition of
  facial attributes.
\newblock In {\em European Conference on Computer Vision (ECCV)}, pages 19--35.
  Springer, 2016.

\bibitem{schuckers2022statistical}
M.~Schuckers, S.~Purnapatra, K.~Fatima, D.~Hou, and S.~Schuckers.
\newblock Statistical methods for assessing differences in false non-match
  rates across demographic groups.
\newblock In {\em International Conference on Pattern Recognition (ICPR)},
  pages 570--581. Springer, 2022.

\bibitem{serna2022sensitive}
I.~Serna, A.~Morales, J.~Fierrez, and N.~Obradovich.
\newblock Sensitive loss: Improving accuracy and fairness of face
  representations with discrimination-aware deep learning.
\newblock {\em Artificial Intelligence}, 2022.

\bibitem{terhorst2020post}
P.~Terh{\"o}rst, J.~N. Kolf, N.~Damer, F.~Kirchbuchner, and A.~Kuijper.
\newblock Post-comparison mitigation of demographic bias in face recognition
  using fair score normalization.
\newblock {\em Pattern Recognition Letters}, 140:332--338, 2020.

\bibitem{terhoerst2020comparison}
P.~Terh{\"o}rst, M.~L. Tran, N.~Damer, F.~Kirchbuchner, and A.~Kuijper.
\newblock Comparison-level mitigation of ethnic bias in face recognition.
\newblock In {\em International Workshop on Biometrics and Forensics (IWBF)}.
  IEEE, 2020.

\bibitem{vangara2019characterizing}
K.~Vangara, M.~C. King, V.~Albiero, K.~Bowyer, et~al.
\newblock Characterizing the variability in face recognition accuracy relative
  to race.
\newblock In {\em Conference on Computer Vision and Pattern Recognition (CVPR)
  Workshops}, 2019.

\bibitem{wallace2012cross}
R.~Wallace, M.~McLaren, C.~McCool, and S.~Marcel.
\newblock Cross-pollination of normalisation techniques from speaker to face
  authentication using gaussian mixture models.
\newblock {\em Transactions on Information Forensics and Security (TIFS)},
  7(2), 2012.

\bibitem{wang2018cosface}
H.~Wang, Y.~Wang, Z.~Zhou, X.~Ji, D.~Gong, J.~Zhou, Z.~Li, and W.~Liu.
\newblock {CosFace}: Large margin cosine loss for deep face recognition.
\newblock In {\em Conference on Computer Vision and Pattern Recognition
  (CVPR)}, 2018.

\bibitem{wang2020mitigating}
M.~Wang and W.~Deng.
\newblock Mitigating bias in face recognition using skewness-aware
  reinforcement learning.
\newblock In {\em Conference on Computer Vision and Pattern Recognition
  (CVPR)}, 2020.

\bibitem{wang2021survey}
M.~Wang and W.~Deng.
\newblock Deep face recognition: A survey.
\newblock {\em Neurocomputing}, 429:215--244, 2021.

\bibitem{wang2019racial}
M.~Wang, W.~Deng, J.~Hu, X.~Tao, and Y.~Huang.
\newblock Racial faces in the wild: Reducing racial bias by information
  maximization adaptation network.
\newblock In {\em International Conference on Computer Vision (ICCV)}, pages
  692--702. IEEE, 2019.

\bibitem{wang2019deep}
P.~Wang, F.~Su, Z.~Zhao, Y.~Guo, Y.~Zhao, and B.~Zhuang.
\newblock Deep class-skewed learning for face recognition.
\newblock {\em Neurocomputing}, 2019.

\bibitem{xu2021consistent}
X.~Xu, Y.~Huang, P.~Shen, S.~Li, J.~Li, F.~Huang, Y.~Li, and Z.~Cui.
\newblock Consistent instance false positive improves fairness in face
  recognition.
\newblock In {\em Conference on Computer Vision and Pattern Recognition
  (CVPR)}, pages 578--586, 2021.

\bibitem{yang2022enhancing}
Y.~Yang, A.~Gupta, J.~Feng, P.~Singhal, V.~Yadav, Y.~Wu, P.~Natarajan,
  V.~Hedau, and J.~Joo.
\newblock Enhancing fairness in face detection in computer vision systems by
  demographic bias mitigation.
\newblock In {\em AAAI/ACM Conference on AI, Ethics, and Society}, 2022.

\bibitem{yang2021ramface}
Z.~Yang, X.~Zhu, C.~Jiang, W.~Liu, and L.~Shen.
\newblock {RamFace}: race adaptive margin based face recognition for racial
  bias mitigation.
\newblock In {\em International Joint Conference on Biometrics (IJCB)}. IEEE,
  2021.

\bibitem{zheng2018cross}
T.~Zheng and W.~Deng.
\newblock Cross-pose {LFW}: A database for studying cross-pose face recognition
  in unconstrained environments.
\newblock Technical report, Beijing University of Posts and Telecommunications,
  2018.

\bibitem{zhu2021webface260m}
Z.~Zhu, G.~Huang, J.~Deng, Y.~Ye, J.~Huang, X.~Chen, J.~Zhu, T.~Yang, J.~Lu,
  D.~Du, et~al.
\newblock {WebFace260M}: A benchmark unveiling the power of million-scale deep
  face recognition.
\newblock In {\em Conference on Computer Vision and Pattern Recognition
  (CVPR)}, 2021.

\end{thebibliography}
